\documentclass{article}
\usepackage[letterpaper,top=2cm,bottom=2cm,left=3cm,right=3cm,marginparwidth=1.75cm]{geometry}
\usepackage{graphicx} 
\usepackage{soul}

\newcommand\blfootnote[1]{%
  \begingroup
  \renewcommand\thefootnote{}\footnote{#1}%
  \addtocounter{footnote}{-1}%
  \endgroup
}
\newcommand{\LMM}{MFM}

\usepackage{url}
\usepackage{alltt}
\usepackage[most]{tcolorbox}
\tcbset{
  aibox/.style={
    width=474.18663pt,
    top=10pt,
    colback=white,
    colframe=black,
    colbacktitle=black,
    enhanced,
    center,
    attach boxed title to top left={yshift=-0.1in,xshift=0.15in},
    boxed title style={boxrule=0pt,colframe=white,},
  }
}
\newtcolorbox{AIbox}[2][]{aibox,title=#2,#1}

\title{\st{Solving Robotics with GPTs} \\ Towards Generalist Robots: Scaling Up Low-Level Motor Skill Learning via Automated Task and Scene Generation}
\title{(Potentially) The Final Piece towards Generalist Robots: Large-Scale Low-Level Motor Skill Learning via Automated Task and Scene Generation}
\title{Towards A Foundation Model for Generalist Robots: \\Diverse Skill Learning at Scale via Automated Task and \\ Scene Generation}

\title{Towards Generalist Robots: \\A Promising Paradigm via \textit{Generative Simulation}}

\author{\normalsize Zhou Xian$^{1*}$, Theophile Gervet$^1$,  Zhenjia Xu$^2$, \\\normalsize Yi-Ling Qiao$^3$, Tsun-Hsuan Wang$^4$, Yian Wang$^5$, Yufei Wang$^1$}
\date{\normalsize $^1$CMU \quad
$^2$Columbia University \quad
$^3$University of Maryland \quad
$^4$MIT \quad
$^5$UMass Amherst \\ \large \vspace{1em} Version 0.3, August 29th, 2023 \\\vspace{0.2em} \small (Initial release: May 16th, 2023)}

\begin{document}

\maketitle
\begin{abstract}
This document serves as a position paper that outlines the authors' vision for a potential pathway towards \textit{generalist robots}. The purpose of this document is to share the excitement of the authors with the community and highlight a promising research direction in robotics and AI.
The authors believe the proposed paradigm is a feasible path towards accomplishing the long-standing goal of robotics research: deploying robots, or embodied AI agents more broadly, in various \textit{non-factory real-world} settings to perform \textit{diverse} tasks.

This document presents a specific idea for mining knowledge in the latest large-scale foundation models for robotics research. Instead of directly using or adapting these models to produce low-level policies and actions, it advocates for a fully automated generative pipeline (termed as \textit{generative simulation}), which uses these models to generate diversified tasks, scenes and training supervisions at scale, thereby scaling up low-level skill learning and ultimately leading to a foundation model for robotics that empowers generalist robots.


The authors are actively pursuing this direction, but in the meantime, they recognize that the ambitious goal of building generalist robots with large-scale policy training demands significant resources such as computing power and hardware, and research groups in academia alone may face severe resource constraints in implementing the entire vision. Therefore, the authors believe sharing their thoughts at this early stage could foster discussions, attract interest towards the proposed pathway and related topics from industry groups, and potentially spur significant technical advancements in the field.


\blfootnote{$^*$Corresponding author: zhouxian@cmu.edu}

\end{abstract}

\section{The Question}

Recent years, and especially recent months (Yes, technology is iterating at an unprecedented monthly-based frequency today. How crazy is that?), have witnessed tremendous progress in multimodal foundation and generative models. These models have achieved remarkable performance in generating texts \cite{vaswani2017attention, devlin2018bert, brown2020gpt3, touvron2023llama, alpaca, driess2023palm, ChatGPT}, images \cite{rombach2022high, ramesh2022hierarchical, kang2023scaling, saharia2022photorealistic, girdhar2023imagebind}, 3D shapes and scenes \cite{zeng2022lion, poole2022dreamfusion, saito2020pifuhd, liu2023meshdiffusion, jun2023shap, melas2023realfusion, nichol2022point}, audios \cite{ghosal2023text, liu2023audioldm, huang2023make} and videos \cite{singer2022make, khachatryan2023text2video, luo2023videofusion}. Notably, with the recent release of several multimodal fundation models including GPT-4 \cite{gpt4} and others \cite{zhu2023minigpt, gong2023multimodal, li2023otter, girdhar2023imagebind}, the performance and usability of these models have finally surpassed an inflection point, making them almost fully functional as a universal interface that can interact with a wide range of different tools \cite{plugin, schick2023toolformer, suris2023vipergpt, yang2023mm, shen2023hugginggpt}, thereby solving diverse tasks spanning mathematics, coding, vision, medicine, law, psychology, and more \cite{bubeck2023sparks} for people in the \textit{virtual world}.
But what about in the \textit{physical world}? Despite the impressive progress in these foundation models, their capabilities are still confined to operating within the realms of virtual data and language space. In order to interact with the physical world, \textit{robots}, or \textit{embodied agents}, must not only perceive, reason, and plan using vision and language but also execute physical actions and interact with their environments. Unfortunately, existing foundation models have been trained primarily on \textit{virtual data} available on the internet and have yet to incorporate physical action capabilities. As a result, a considerable gap remains between the current capabilities of these models and the requirements for robots to effectively perform a wide range of skills in the physical world.

For the past decades, directions in robotics research can be broadly grouped into two major families: 1) improving robots' high-level cognition, reasoning and planning capabilities (e.g., recognizing objects, understanding instructions, generating sub-goals or sub-tasks given a high-level goal, planning path for navigation, etc.) \cite{gervet2022navigating, ahn2022can, shridhar2022cliport, zeng2021transporter, gkanatsios2023energy, tung20203d, zhang2021world} and 2) enabling robots with low-level motor skills (e.g., locomotion skills for humanoids or legged robots, dexterous manipulation skills, tool using, etc.) \cite{suarez2018can, xian2017closed, xu2023roboninja, xian2023fluidlab, wang2023softzoo, radosavovic2023learning, xian2021hyperdynamics, chi2023diffusion, lin2022diffskill}. Lucky for (and somewhat surprising to) robotics researchers, recent advancements in foundation models have already demonstrated impressive capabilities for cognition and planning \cite{bubeck2023sparks, huang2022inner, liang2022code, ahn2022can, radford2021learning, gpt4}. Though it would not be entirely accurate to claim that these high-level problems are completely solved, the capabilities provided by the large models have already proven useful in a wide range of task settings \cite{chen2022open, stone2023open, yu2023scaling, driess2023palm}. 
We suggest that, \textit{potentially}, the last missing piece towards generalist robotic agents may lie in equipping them with a \textit{diverse} range of low-level motor skills, which can be seamlessly activated in response to specific sub-tasks generated by high-level planners given a high-level task or goal.

The recurring and significant successes we have witnessed in the past achieved by simply scaling up language and vision models, coupled with architectures with minimal yet proper inductive biases \cite{trockman2022patches, tolstikhin2021mlp, vaswani2017attention}, indicate that leveraging ever-increasing computational power through the integration of both large-scale models and large-scale data may be the simplest, yet most promising and effective approach across various fields \cite{sutton2019bitter}. We believe this also applies to the field of robotics, or more specifically, the direction of acquiring diverse low-level motor skills in a \textit{unified} way. Given that attention-based architecture is thriving in various areas including robotics \cite{radosavovic2023learning, shridhar2023perceiver, liu2022instruction, guhur2023instruction, brohan2022rt}, probably the \textit{most important question} that remains is: 
\bigbreak
\textit{How do we \textbf{scale up data collection} for robotics to the level that matches the data scale used in training existing large language and vision models, thereby enabling robots to master a \textit{variety} of low-level motor skills across \textit{diverse} task settings with a \textbf{unified neural policy}? Additionally, if existing multimodal foundation models can accelerate this process by leveraging the \textbf{virtual} yet \textbf{general knowledge} they encode, what is the right way to use them?}

\section{The Idea}
\subsection{Why}
Following the rapid advancement in large model training, many ongoing research works in robotics is making efforts to leverage large multimodal models to generate robotics policies via different manners, such as code generation \cite{liang2022code, wu2023tidybot}, data augmentation via visual imagination \cite{yu2023scaling, dai2023learning}, sub-task planning \cite{huang2022inner, ahn2022can, lin2023text2motion, kapelyukh2023dall}. However, attempts to this day only demonstrate simplistic task settings, which can mostly be framed as pick-and-place (discrete end-effector pose generation) following \cite{zeng2021transporter, james2022q, shridhar2022cliport}, and have yet to show promise to scale up to \textit{diverse motor skills}.
Other works strive to scale up robotic data collection in the real world via either kinesthetic teaching or teleoperation \cite{mandlekar2021matters, brohan2022rt, chi2023diffusion, cui2022play}, or leveraging visual human demonstration \cite{young2021visual, sieb2020graph, wang2023mimicplay}. However, we argue data collection in real-world might not be the optimal direction, as it suffers from the following issues:

\begin{itemize}

    \item \textit{Active kinesthetic} demonstrations necessitate considerable investment in robotic hardware and human effort. Human operators must operate robots physically in the real world, and kinesthetic teaching or teleoperation requires operators to undergo additional training. These factors render this approach unsuitable for scaling up data collection to the desired level. For illustration, researchers who built RT-1 \cite{brohan2022rt} spent 17 months using a fleet of 13 robots to gather demonstration data, ultimately yielding a model capable of executing only basic pick-and-place tasks with reliable generalization to new environments left as an open problem.

    \item Conversely, \textit{passive visual} human demonstrations are more cost-effective to collect — and can be found at scale on the internet — but tend to be unstructured and noisy. Recent progress along this thread showcases the potential to extract useful information from passive human videos for manipulation, such as object affordances \cite{mendonca2023alan, ye2023affordance}, ways of hand-object interaction \cite{patel2022learning, bahl2022human}, and goal-conditioned motion trajectories \cite{liu2022joint, wang2023mimicplay}. However, due to the embodiment gap between humans and robotic arms and end-effectors, mapping these information to robot configurations for fine-grained control remains challenging \cite{young2021visual, sieb2020graph, pari2021surprising}.

    \item Human demonstrations typically offer only spatial trajectory information, which are inadequate for tasks beyond basic pick-and-place operations. For instance, even seemingly simple tasks, such as wiping a table or opening a soda can, demand high-frequency closed-loop control of the robot's end-effector conditioned on real-time visual and tactile feedback. Naive mimicking human-provided spatial trajectories struggles to recover from cases that fall outside the expected data distribution.
\end{itemize}

Considering the challenges involved in data collection in real-world environments, we contend that \textit{large-scale simulation} holds considerable potential as an alternative approach. Simulation affords robotic agents unlimited trial-and-error and multimodal sensory inputs \cite{xu2023roboninja, xu2023efficient, si2022taxim, zhang2023close}, offers accelerated and parallel computation for faster data generation, and allows sidestepping potential risks and dangers associated with real-world experiments.
By training in simulation, robots have already acquired a wide range of challenging skills, including but not limited to: multi-finger manipulation \cite{chen2022system, chen2022towards}, assembling furniture pieces \cite{suarez2018can}, deformable object manipulation \cite{lin2022diffskill, huang2021plasticinelab, xu2023roboninja, xian2023fluidlab}, acrobatic flight \cite{kaufmann2020deep, loquercio2021learning}, and locomotion skills for legged robots \cite{fu2022coupling, kumar2021rma}, humanoid robots \cite{radosavovic2023learning}, and soft robots \cite{wang2023softzoo}.
Sim-to-real transfer techniques like domain randomization \cite{tobin2017domain, radosavovic2023learning, che2019domain, haarnoja2023learning, peng2018sim}, modularity and abstraction \cite{muller2018driving, loquercio2021learning}, history encoding \cite{kumar2021rma, fu2022coupling} and realistic sensory signal simulation \cite{zhang2023close, zheng2022luisarender, mildenhall2021nerf} have already enabled transferring tasks as complex and dynamical as humanoid locomotion \cite{radosavovic2023learning} and soccer playing \cite{haarnoja2023learning} from simulation to the real world.

However, simulation-trained policies are usually limited to specific tasks and short horizons. We contend that the primary reasons for the limited generality of simulation-trained policies are twofold: 1) RL or trajectory optimization is effective for generating short-horizon task-specific low-level motor skills, but effectively extending them to long-horizon tasks involving sub-goal planning and skill switching still remains as an open research problem; and 2) there lacks a mechanism to scale up the diversity of tasks and scenes in simulation. However, this is \textit{precisely} where the latest multimodal foundation models (\LMM s\footnote{We will be using the term \LMM~to loosely refer to existing large language models, image generative models (GAN- or diffusion-based), multimodal models (such as GPT-4 \cite{gpt4} or ImageBind \cite{girdhar2023imagebind}), and generative models supporting other modalities.}) can help! 
 We recognize that 1) existing \LMM s can help decompose a high level task into sub-tasks with arbitrarily fine granularity \cite{ahn2022can, liang2022code, driess2023palm, bubeck2023sparks}, which allows control policies to \textit{concentrate} on accomplishing \textit{low-level sub-tasks}, and 2) they can further help significantly scale up the diversity of task and scene configurations in simulation, via automated \textit{task} and \textit{scene generation} in simulation.
 
Today's \LMM s have already demonstrated exceptional capabilities in generating data across various modalities, including
texts \cite{vaswani2017attention, devlin2018bert, brown2020gpt3, touvron2023llama, alpaca, driess2023palm, ChatGPT}, images \cite{rombach2022high, ramesh2022hierarchical, kang2023scaling, saharia2022photorealistic, girdhar2023imagebind}, 3D shapes and scenes \cite{liu2023zero, zeng2022lion, poole2022dreamfusion, saito2020pifuhd, liu2023meshdiffusion, jun2023shap, melas2023realfusion, nichol2022point}, and code \cite{wang2021codet5, ChatGPT, li2022competition, liang2022code}. Leveraging the current generative capacities of these models, we propose further generating diverse \textit{low-level sub-task descriptions}, and subsequently \textit{task-specific scene components}, \textit{configurations}, and training supervisions such as \textit{reward} or \textit{loss functions}.
Conditioned on such sub-task level descriptions, an \LMM~can further generate tasks and scenes (e.g., via code or configuration file generation) to reconstruct desired environments in simulation. 

Unlike prior methods such as procedural generation, manual labelling, program synthesis, or even random sampling \cite{deitke2022procthor, li2023behavior, dalal2021accelerating}, tasks and scenes generated by \LMM s have the potential to closely resemble the \textit{distribution of real-world scenarios}. By processing all human-generated internet language and image corpora, these models have developed comprehensive knowledge of data distributions of the world, which we can harness to get simulations much closer to the full breadth of the real world. 
Once a diverse set of scene configurations and their corresponding fine-grained low-level tasks are generated, we can proceed to delegate the task of policy learning to established RL or trajectory optimization methods, which have demonstrated efficacy in acquiring low-level, short-horizon skills across various applications \cite{chen2022towards, lin2022diffskill, huang2021plasticinelab, xu2023roboninja, xian2023fluidlab, kaufmann2020deep, loquercio2021learning, kumar2021rma}.


In summary, our response to the question is: 

\bigbreak
\textit{We should exploit the full potential of the latest \LMM s to the \textbf{greatest} extent of their capabilities, but \textbf{not} further, and leave the remaining tasks that rely on knowledge of actual physical interactions to \textbf{large-scale skill training} within simulations using \textbf{auto-generated tasks}, \textbf{scenes} and training supervisions such as \textbf{reward functions}}.

\subsection{How}
\begin{figure*}[t!]
\begin{AIbox}{}
{\bf Prompt:} 

Suppose you are a household robot. I want you to clean up my kitchen. What are the specific tasks you would do? Give me 5 examples without details. \\

\parbox[t]{\textwidth}
{{\bf GPT-4:} \begin{alltt}
1. Washing the dishes.\\
2. Cleaning the countertops.\\
3. Sweeping the floor.\\
4. Emptying the trash.\\
5. Organizing the pantry.
\end{alltt}}
\vspace{5pt}

{\bf Prompt:} 

I would like to proceed with ``sweeping the floor''. Could you give me step-by-step instructions to complete the task? Only short description for each step is needed.\\

\parbox[t]{\textwidth}
{{\bf GPT-4:} \begin{alltt}
1. Locate broom and dustpan. \\
2. Grasp broom handle.\\
3. Begin sweeping from one corner of the room.\\
4. Sweep debris towards dustpan.\\
5. Hold dustpan in place.\\
6. Sweep debris into dustpan.\\
7. Dispose of debris in trash bin.\\
8. Return broom and dustpan to storage.\\
\end{alltt}}
\end{AIbox}
\caption{Generating sub-tasks given a high-level goal.}
\label{fig:subtask}
\end{figure*}

\begin{figure*}[t!]
\begin{AIbox}{}
{\bf Prompt:} 

Imagine you are about to execute the step ``grasp broom handle''. Before you execute it, could you imagine the current environment you are in?
Provide some typical objects and components in the current scene with just the names. Make sure to include necessary components for the task you are about to perform.\\

\parbox[t]{\textwidth}
{{\bf GPT-4:} \begin{alltt}
1. Kitchen floor\\
2. Cabinets\\
3. Broom\\
4. Dustpan\\
5. Trash bin\\
6. Countertops\\
7. Refrigerator\\
8. Stove\\
9. Sink\\
\end{alltt}}
\end{AIbox}
\caption{Generating scene components given a sub-task.}
\label{fig:component}
\end{figure*}

\begin{figure*}[t!]
\begin{AIbox}{}
{\bf Prompt:} 

I am trying to generate such a scene in simulation. Could you tell me how should I place them? Imagine we are in a Cartesian space, where x-y is the horizontal plane and z is the height, what would the objects you just listed be placed at typically? \\

\parbox[t]{\textwidth}
{{\bf GPT-4:} \begin{alltt}
Assuming the kitchen floor is a 4m x 4m square, and the origin (0m, 0m, 0m) is at the bottom left corner, objects could be placed as follows:\\
1. Kitchen floor: Already placed at (0m, 0m, 0m)\\
2. Cabinets: Along the walls at (0.5m, 0m, 0.9m) and (0m, 0.5m, 0.9 m)\\
3. Broom: Leaning against the wall at (3.5m, 3m, 0m)\\
4. Dustpan: Hanging on the wall next to the broom at (3.5m, 3.2m, 0m)\\
5. Trash bin: Positioned at (3m, 3m, 0m)\\
6. Countertops: Along the walls at (0m, 0m, 0.9m) and (0m, 1.5m, 0.9m)\\
7. Refrigerator: Located at (0m, 3m, 0m)\\
8. Stove: Positioned at (2m, 0m, 0m)\\
9. Sink: Situated in the countertop at (1m, 0m, 0.9m)

\end{alltt}}
\vspace{5pt}

{\bf Prompt:} 

My simulator follows this API: you can add an object A at (x, y, z) by calling function \texttt{sim.add\_object(A, pos=(x, y, z))}. Translate the above configuration to code. Include walls and floors.\\

\parbox[t]{\textwidth}
{{\bf GPT-4:} \begin{alltt}
sim.add\_object("Kitchen Floor", pos=(0, 0, 0)) \\
sim.add\_object("Wall 1", pos=(0, 0, 0)) \\
sim.add\_object("Wall 2", pos=(0, 0, 0)) \\ sim.add\_object("Wall 3", pos=(4, 0, 0)) \\
sim.add\_object("Wall 4", pos=(0, 4, 0))\\ sim.add\_object("Cabinets", pos=(0.5, 0, 0.9)) \\
sim.add\_object("Cabinets", pos=(0, 0.5, 0.9)) \\
sim.add\_object("Broom", pos=(3.5, 3, 0)) \\
sim.add\_object("Dustpan", pos=(3.5, 3.2, 0)) \\
sim.add\_object("Trash Bin", pos=(3, 3, 0)) \\
sim.add\_object("Countertops", pos=(0, 0, 0.9)) \\
sim.add\_object("Countertops", pos=(0, 1.5, 0.9)) \\
sim.add\_object("Refrigerator", pos=(0, 3, 0)) \\
sim.add\_object("Stove", pos= (2, 0, 0))\\
sim.add\_object("Sink", pos= (1, 0, 0.9))
\end{alltt}}
\end{AIbox}
\caption{Generating scene configurations.}
\label{fig:scene}
\end{figure*}

\begin{figure*}[t!]
\begin{AIbox}{}
{\bf Prompt:} 

I want to you to design a reward function for me for training a robotic agent performing the task ``grasping the broom''. \\

\parbox[t]{\textwidth}
{{\bf GPT-4:} \\
Designing a reward function for a robotic agent to perform the task ``grasping the broom'' requires considering several factors like the position of the gripper, the distance between the gripper and the broom, and whether the broom is between the gripper fingers. Here's an example reward function for this task:

\begin{alltt}
def reward function(sim):\\
\hspace*{10pt}\# Constants \\
\hspace*{10pt}GRIPPER\_DISTANCE THRESHOLD = 0.01\\
\hspace*{10pt}BROOM\_RADIUS = 0.02\\

\hspace*{10pt}\# Get the position of the gripper, broom, and gripper fingers\\
\hspace*{10pt}gripper\_position = sim.robot.gripper.position\\
\hspace*{10pt}broom\_position = sim.environment.broom.position\\

\hspace*{10pt}left\_finger\_position = sim.robot.gripper.left\_finger.position\\
\hspace*{10pt}right\_finger\_position = sim.robot.gripper.right\_finger.position\\

\hspace*{10pt}\# Calculate the distance between the gripper and the broom\\
\hspace*{10pt}distance = np.linalg.norm(gripper\_position - broom\_position)\\

\hspace*{10pt}\# Determine if the broom is between the gripper fingers\\
\hspace*{10pt}is\_between = is\_between\_points(broom\_position, left\_finger\_position, right\_finger\_position)\\

\hspace*{10pt}\# Calculate the reward\\
\hspace*{10pt}if distance < GRIPPER\_DISTANCE\_THRESHOLD and is\_between:\\
\hspace*{20pt}reward = 1\\
\hspace*{10pt}else:\\
\hspace*{20pt}reward = -distance / (BROOM\_RADIUS + GRIPPER DISTANCE THRESHOLD)\\
\hspace*{10pt}return reward\\

def is\_between\_points(target\_point, point\_A, point\_B):\\
\hspace*{10pt}\# Find the vectors between target\_point and point\_A, target\_point and point\_B\\
\hspace*{10pt}vec\_target\_A = point\_A - target\_point\\
\hspace*{10pt}vec\_target\_B = point\_B - target\_point\\

\hspace*{10pt}\# Calculate the dot product between these two vectors\\
\hspace*{10pt}dot\_product = np.dot(vec\_target\_A, vec\_target\_B)\\

\hspace*{10pt}\# If the dot product is less than or equal to zero, the target point is between point A and point B\\

\hspace*{10pt}return dot\_product <= 0\\
\end{alltt}}
\vspace{5pt}

\end{AIbox}
\caption{Generating a task-specific reward function for ``grasping the broom''.}
\label{fig:reward1}
\end{figure*}

\begin{figure*}[t!]
\begin{AIbox}{}
{\bf Prompt:} 

Similarly, design a reward for the task ``sweep the floor''. \\

\parbox[t]{\textwidth}
{{\bf GPT-4:} \\
To design a reward function for the task ``sweep the floor'', we need to consider several factors such as the broom's position, orientation, contact with the debris, and the progress of moving debris towards the target location. Here's an example reward function for this task:

\begin{alltt}
def reward\_function(sim):\\
\hspace*{10pt}SWEEPING\_DISTANCE\_THRESHOLD = 0.1\\
\hspace*{10pt}TARGET\_REACHED\_THRESHOLD = 0.2\\
\hspace*{10pt}BROOM\_ORIENTATION\_THRESHOLD = 0.1\\

\hspace*{10pt}broom\_position = sim.environment.broom.position\\
\hspace*{10pt}broom\_orientation = sim.environment.broom.orientation\\
\hspace*{10pt}debris\_positions = [debris.position for debris in sim.environment.debris\_list]\\
\hspace*{10pt}target\_position = np.array(target)\\
\hspace*{10pt}broom\_aligned = is\_broom\_aligned(broom\_orientation, BROOM\_ORIENTATION\_THRESHOLD)\\
\hspace*{10pt}debris\_distances = [np.linalg.norm(broom\_position - debris\_position) for debris\_position in debris\_positions]\\
\hspace*{10pt}close\_debris\_indices = [index for index, distance in enumerate(debris\_distances)\\

\hspace*{10pt}if distance < SWEEPING\_DISTANCE\_THRESHOLD]\\
\hspace*{10pt}\# Calculate the progress of moving debris towards the target location\\
\hspace*{10pt}progress = 0\\
\hspace*{10pt}for index in close\_debris\_indices:\\
\hspace*{10pt}\hspace*{10pt}debris\_position = debris\_positions[index]\\
\hspace*{10pt}\hspace*{10pt}initial\_distance = np.linalq.norm(debris position - target position)\\
\hspace*{10pt}\hspace*{10pt}updated\_distance = np.linalg.norm(sim.environment.debris\_list[index].updated\_position - target position)\\
\hspace*{10pt}\hspace*{10pt}if updated distance < initial distance:\\
\hspace*{10pt}\hspace*{10pt}\hspace*{10pt}progress += (initial\_distance - updated\_distance)\\

\hspace*{10pt}\# Calculate the reward based on broom alignment and progress towards the target\\
\hspace*{10pt}reward = 0\\
\hspace*{10pt}if broom\_aligned:\\
\hspace*{10pt}\hspace*{10pt}reward += 0.5\\
\hspace*{10pt}\hspace*{10pt}reward + = progress / (TARGET\_REACHED\_THRESHOLD * len(close\_debris\_indices))\\
\hspace*{10pt}return reward\\

def is\_broom\_aligned(broom\_orientation, threshold):\\
\hspace*{10pt}\# Assuming the broom should be parallel to the ground (aligned with the x-y plane)\\
\hspace*{10pt}\# The angle between the broom's orientation and the z-axis should be close to 90 degrees\\
\hspace*{10pt}broom\_orientation = broom\_orientation / np.linalg.norm(broom\_orientation)\\
\hspace*{10pt}z\_axis = np.array([0, 0, 1])\\
\hspace*{10pt}dot\_product = np.dot(broom \_orientation, z axis)\\
\hspace*{10pt}angle = np.arccos(dot product) * (180 / np.pi)\\
\hspace*{10pt}return abs(angle - 90) < threshold\\
\end{alltt}}
\vspace{5pt}

\end{AIbox}
\caption{Generating a task-specific reward function for ``sweep the floor''.}
\label{fig:reward2}
\end{figure*}

The most appealing aspect of this paradigm is that every component required to automate and scale up task and scene generation falls neatly within the capabilities and modalities of today's \LMM s, and can be accomplished by prompting them with proper instructions and in-context examples. We call this paradigm \textit{generative simulation}, which can be used for scaling up robot learning in fully automated way with minimal human involvement. To provide a more comprehensive explanation, we sketch a possible pipeline with the following stages:
\begin{enumerate}
    \item \textbf{Sub-task}
    \begin{enumerate}
        \item \textbf{Given high-level goals:} Given a high-level goal (e.g. ``clean up a kitchen"), we can query an \LMM~to generate step-by-step sub-tasks. Figure \ref{fig:subtask} presents a simple example using GPT-4 to illustrate this process. This enables the generation of fine-grained sub-task descriptions by recursively querying an \LMM. Furthermore, we can even automate the generation of numerous high-level goals by initiating the conversation with e.g. ``\textit{What are the top 10 desired tasks a household robot should perform for humans?}'', and then \textit{progressively} and \textit{iteratively} delving into finer-granular levels until we reach the desired task granularity. 
        \item \textbf{Given scene configurations:} Apart from conditioning on a high-level goal, \LMM s also possess the capacity to generate potential sub-tasks in given a current scene configuration. For example, given a kitchen scenario (either described in natural language suited for language models, or depicted in an image suited for models such as GPT-4 that are capable of handling image input), an \LMM~could generate an array of sub-tasks taking into account \textit{affordances} of the objects in the scene, such as ``open the middle-drawer of a cabinet'' or ``turn the oven knob''. This could substantially speed up exploration and skill acquisitions. 
        
    \end{enumerate}

    \item \textbf{Scene component}\\ If a sub-task is generated from a high-level goal, we can next construct a corresponding simulation scene for training this sub-task. This process begins with populating the scene with plausible and common objects and components, which can be readily acquired by querying \LMM s using proper instructions. See Figure \ref{fig:component} for an example.
    \item \textbf{Scene configuration} 
    \begin{enumerate}
        \item \textbf{Bottom-up (in language space):} Once we have obtained necessary components for generating the scene, we can query the \LMM~to produce the spatial configurations of these components, and subsequently translate them into code (or \texttt{JSON}-like configuration files and other possible formats) that can be executed by simulators. Figure \ref{fig:scene} illustrates how this can be done nicely via simply querying GPT-4; Note that we haven't yet spent any effort on prompt engineering or providing in-context examples. Unlike random sampling spatial configurations of the objects, which could result in unrealistic scenarios such as a water bottle floating in the air, this process yields spatial configurations resembling real-world distributions. Furthermore, if specifying spatial configurations in Cartesian coordinate doesn't produce plausible configurations as desired, we can build such preliminary configuration in simulations first, render the visual appearances, and query \LMM s iteratively with rendered views to determine if they appear plausible. 
        \item \textbf{Top-down (in image space):} Another alternative for scene generation is top-down: we can first generate a desired scene as a whole in image space (either from scratch or conditioned on the components generated above), and then detect the generated scene components and reconstruct it in simulation. It would require empirical experiments to determine whether the bottom-up or top-down approach works better in practice.
    \end{enumerate}
    \item \textbf{Textures, object, and dynamics}\\
    Before populating the scene with the generated object and component choices, we need also the actual object assets for both rendering and simulation. There are multiple possible ways for obtaining object meshes and textures: we can either directly query 3D content generative models for object, scene and texture generation \cite{zeng2022lion, poole2022dreamfusion, saito2020pifuhd, liu2023meshdiffusion, jun2023shap, nichol2022point, hollein2023text2room, gupta20233dgen, chen2023text2tex} to generate them, or additionally provide some example assets to bootstrap on \cite{liu2023zero, melas2023realfusion}, using existing mesh and texture datasets \cite{xiang2020sapien, deitke2022objaverse}. We suspect the latter would work better for complex cases such as articulated objects. Once we have the necessary object assets, we can supply the \LMM s with all the material models that the simulator supports (e.g., rigid objects, elastic objects, plastic objects, viscous fluids, inviscid fluids, gas, etc. \cite{xian2023fluidlab}) and query for the most appropriate model and physical properties for each corresponding object. 
    \item \textbf{Reward or loss functions}\\
    The final stage involves generating the reward function, conditioned on the current sub-task and scene. We hypothesize that, once a high-level task is decomposed into the finest-grained sub-tasks, long-horizon training is no longer necessary, and reward or loss functions for such fine-grained sub-tasks can be represented in a simple form, again falling within the modalities supported by \LMM s. One straightforward approach to achieve this is by prompting \LMM s with example reward definitions and requesting it to generate more based on the current scenario. Figure \ref{fig:reward1} and  \ref{fig:reward2} illustrate two examples of this process. It is important to note that the second ``sweep the floor'' task (Figure \ref{fig:reward2}) is still a relatively high-level task and is included here just for illustration purposes, but the \LMM~can still consider multiple relevant factors and generate a reasonable reward definition. In practice, we can further decompose it into even finer sub-routines. Additionally, for tasks whose rewards cannot be represented as simple geometric computations, such as \textit{cooking a steak}, we can generate image-based goal condition and reward functions. Further, to ease the downstream learning of the sub-tasks, \LMM s can be used to design the observation and action space of the RL policy or trajectory optimization algorithms. For example, for the task of ``grasping the broom'', only the location and orientation of the broom and robot end-effector are essential to be included in the policy observation, and all other irrelevant objects such as the debris can be excluded from the observation space. Similarly, for the action space of a trajectory optimizer, \LMM s can specify whether velocity control, position, or force control of the end-effector should be used.

\end{enumerate}

The five stages above collectively form a complete pipeline for the proposed \textit{generative simulation} paradigm, which automatically generates tasks and scenes that can potentially be solved by short-horizon task-specific policy learning. Note that the examples provided are preliminary attempts, and the authors are not any experts in prompt engineering. The example responses in the figures above are generated with \textit{zero} prompt tuning. By leveraging prompt engineering and providing more in-context examples, we expect to achieve responses of significantly higher quality.

Once diverse data for a vast number of individual tasks and configurations are obtained, we can start training a large-scale policy model by either learning from scratch by practicing in the generated scenes, or training task-specific policies first and distill them (e.g. via behavior cloning \cite{chen2022system, xu2023roboninja, lin2022diffskill, reed2022generalist}) into a single generalist policy model conditioned on visual input and language instructions. Appealing advantages of the latter include 1) task-specific policy training can utilize low-level ground-truth system states to accelerate learning, and 2) we can use the converged performance of the task-specific policies as a metric, and \textit{filter out} unreasonable sub-tasks and noisy learned task-specific policies. Furthermore, how to fuse such knowledge of physical interactions into existing \LMM s is a promising and exciting direction to explore. This fusion process would result in a large-scale foundation model that additionally supports data modality for understanding low-level physical interaction and control, paving the way towards a \textbf{generalist embodied AI}. We believe this unified approach also opens up new possibilities, such as few-shot visual imitation via in-context learning for robotics: by leveraging the foundational knowledge of physical actions acquired through such large-scale training and fusing this new mode of knowledge into existing \LMM s, robotic agents can potentially perform few-shot imitation learning for acquiring new skills conditioned on human demonstrations provided visually, as it would be able to draw connections between different modalities and build a core understanding of the physical world from an \textit{embodied} perspective.

\subsection{Limitations}

The authors acknowledge that the proposed paradigm makes several assumptions and has the following limitations:
\begin{enumerate}
    \item The proposed approach relies on the assumption that a fully capable and unified simulation platform is available, supporting 1) simulation of diverse materials and objects, as well as diverse physics phenomena, and 2) realistic sensory feedback including RGB, depth, tactile, force-torque, IMU, acoustic signal and more. In recent years, numerous works have been proposed and developed in this direction \cite{xian2023fluidlab, huang2021plasticinelab, radosavovic2023learning, xu2023roboninja, Omniverse, li2023behavior, si2022taxim}. We have also witnessed tremendous progress in realistic simulation techniques for diverse complex materials such as liquid \cite{wu2018fast}, snow \cite{hu2019taichi}, cloth \cite{narain2013folding}, honey \cite{shao2022fast}, bread \cite{ding2019thermomechanical}, meat \cite{wolper2020anisompm}, etc; physical phenomena such as melting \cite{stomakhin2014augmented}, baking \cite{ding2019thermomechanical}, cracking  \cite{pfaff2014adaptive}, etc; as well as physically-based rendering techniques for realistic image \cite{ludvigsen2010real, glassner1989introduction, zheng2022luisarender}, depth rendering \cite{zhang2023close} and tactile signal rendering \cite{si2022taxim, xu2023efficient}. These advancements are making simulated environments increasingly closer to the real world. Although a fully unified simulation platform is still not ready today, it is arguable that the development of such a platform is within our reach. In fact, even if simulating physical phenomena is difficult for complex tasks like cooking, simply applying the proposed approach to relatively simple task settings like object grasping and pick-and-place, will dramatically reduce the cost for generating task and scene variations.
    \item The proposed approach is based on the premise that short-horizon low-level motor skills can be trained effectively, a conclusion reached by the authors in light of recent advancements in various task-specific research \cite{xian2023fluidlab, huang2021plasticinelab, xu2023roboninja, radosavovic2023learning, haarnoja2023learning}. This may not be adequate for solving more complex and highly dynamic tasks, which, even at the lowest granularity, could still be challenging to address through massive parallel reinforcement learning or trajectory optimization. Examples of such tasks include robots performing stunts like skateboard flip tricks or robotic basketball playing. However, we argue that for enabling household robots to perform a wide range of useful tasks for humans, the proposed paradigm is likely sufficient.
    \item The proposed approach assumes that \LMM s can effectively decompose high-level tasks into sub-routines, which might not be sufficient for highly complex tasks requiring intricate reasoning and planning, such as furniture or Lego assembly. In these cases, \LMM s can be prompted with additional information, such as images of instruction manuals or video demonstrations to improve their understanding and provide more accurate and detailed sub-routines, as shown in \cite{wu2023read}.
    \item The proposed approach does not explicitly address how to integrate learning from human demonstrations. However, we posit that focusing on sub-task learning can minimize the reliance on such demonstrations, which also present its own limitations as discussed earlier. Furthermore, for complex low-level tasks where demonstration is essential, we believe the increased diversity in task variations resulting from our approach can also help in reducing cost in constructing diverse scenes for obtaining active demonstration with diverse skills.
    \item The proposed approach is predicated on the assumption that the sim2real can be addressed satisfactorily, in light of a series of recent successful stories of transfer simulation-trained policies to the real world \cite{chen2022system, chen2022towards, lin2022diffskill, xu2023roboninja, kumar2021rma, radosavovic2023learning, radosavovic2023learning, haarnoja2023learning, zhang2023close}. We espouse the view that, provided simulated environments achieve sufficient realism and exhibit a broad spectrum of randomness, the real world essentially constitutes merely a specific instance within this broad spectrum of simulation.

\end{enumerate}

\section{Conclusion}
In this document, we recognize that 1) existing multimodal foundation models (\LMM s) offer both robust high-level cognition and reasoning skills, as well as promising generative capabilities in multiple modalities, and 2) robots can already acquire diverse low-level dexterous skills with task-specific policy learning in simulation. We propose \textit{generative simulation} for robot learning, an automated pipeline to scale up data collection for low-level skill learning in simulation by leveraging \LMM s to automate task and scene generation. Building upon the latest progress in fields spanning computer vision, language models, generative models, policy learning, simulation and rendering, we are excited to witness that for the first time, all the components needed to complete the puzzle towards achieving generalist robotic agents are finally within our reach.

\section*{Acknowledgement}
We thank Chen Wang for help improving the writing of the document. We thank Chuang Gan for coming up with the term \textit{generative simulation}. We would also like to thank all the senior researchers involved in the discussions surrounding this document: Katerina Fragkiadaki, Shuran Song, Chuang Gan, David Held, Chris Atkeson, Ming C. Lin, and Daniela Rus. Their feedback and support have been invaluable in aiding the completeness of the document and providing the encouragement to pursue the proposed line of work.

\bibliographystyle{abbrv}
\bibliography{references}
\end{document}